\pdfoutput=1
\documentclass[11pt]{article}

\usepackage{ACL2023}

\usepackage{times}
\usepackage{latexsym}
\usepackage{hyperref}
\usepackage{gb4e}
\noautomath % Prevents gb4e from interfering with math symbols

\usepackage[T1]{fontenc}
\usepackage[utf8]{inputenc}

\usepackage{microtype}

\usepackage{inconsolata}

\usepackage{color-edits}
\addauthor[Enora]{er}{red}
\addauthor[Alexis]{ap}{cyan}

\title{Interdisciplinary Research in Conversation: A Case Study in Computational Morphology for Language Documentation}

\author{Enora Rice${ }^{1}$ \quad Katharina von der Wense${}^{1,2}$ \quad Alexis Palmer${}^{1}$ \\
    ${ }^{1}$University of Colorado Boulder \quad ${ }^{2}$ Johannes Gutenberg University Mainz  \\ \texttt{enora.rice@colorado.edu}}

\begin{document}
\maketitle
\begin{abstract}
Computational morphology has the potential to support language documentation through tasks like morphological segmentation and the generation of Interlinear Glossed Text (IGT). However, our research outputs have seen limited use in real-world language documentation settings. This position paper situates the disconnect between computational morphology and language documentation within a broader misalignment between research and practice in NLP and argues that the field risks becoming decontextualized and ineffectual without systematic integration of User-Centered Design (UCD). To demonstrate how principles from UCD can reshape the research agenda, we present a case study of GlossLM, a state-of-the-art multilingual IGT generation model. Through a small-scale user study with three documentary linguists, we find that, despite strong metric-based performance, the system fails to meet core usability needs in real documentation contexts. These insights raise new research questions around model constraints, label standardization, segmentation, and personalization.  We argue that centering users not only produces more effective tools, but surfaces richer, more relevant research directions.
\end{abstract}

\section{Introduction}

Morphological analysis plays a central role in language documentation, and computational morphology is well-positioned to support this work through tasks such as morphological segmentation and the generation of Interlinear Glossed Text (IGT), a key linguistic annotation format. Yet, despite over two decades of interest—including early calls for NLP to engage more deeply with endangered languages \citep{bird-2009-last}-- we still lack broadly usable tools that support documentation workflows. This disconnect has been described as the "NLP gap" in language documentation \cite{gessler-2022-closing}, and it presents not only a technical challenge but also a deeper disciplinary mismatch. We add to recent work that has highlighted the importance of incorporating user perspectives and rethinking evaluation practices \citep{ganesh-etal-2023-mind, Liao_Xiao_2025}, and suggest that we need deep structural changes in how interdisciplinary systems are designed and assessed. These changes are especially urgent when research focuses on very low-resource or endangered languages, where care in collaboration is critical: otherwise, we risk building systems that extract data or prestige without meaningfully serving the communities involved \citep{schwartz-2022-primum, bird-2024-must}.

We argue that User-Centered Design (UCD)—an iterative development approach from Human-Computer Interaction that emphasizes early and sustained engagement with end users—offers not only a path to more usable tools for morphological analysis, but also to a richer research process. We illustrate this through a case study of GlossLM \citep{ginn-etal-2024-glosslm}, a state-of-the-art multilingual model for generating IGT. Since the stated aim of \citet{ginn-etal-2024-glosslm} is to "explore the task of automatically generating IGT in order to aid documentation projects," we recruit 3 linguists to complete a small glossing task with GlossLM and share their perspectives on how it might fit into their documentation workflow. Our findings reveal that, despite strong performance on standard metrics, GlossLM falls short for real-world use: it lacks segmentation, enforces prescriptive glossing conventions, and produces out-of-domain labels.  
This feedback enables us to articulate new directions for research that are more accurately grounded in documentation workflows. Our findings raise the following research questions:
\begin{itemize}
    \item[\textbf{Q1.}] Can (and should) we constrain glossing model outputs to pre-defined language specific labels? Or should we instead standardize glossing labels across languages?
    \item[\textbf{Q2.}] Can (and should) we tune glossing model outputs to fit the personal glossing conventions of individual linguists?
    \item[\textbf{Q3.}] Can we do accurate glossing without incorporating declarative language-specific
    information?
    \item[\textbf{Q4.}] Can we extract latent segmentation from glossing models? Do we need to?
\end{itemize}
These questions are not just engineering challenges; they are broader conceptual questions that deserve sustained attention from both computational and linguistic researchers. This case study is just one example of how engaging with users helps surface research directions that are richer, more contextually grounded, and better aligned with the goals of the communities we seek to support. Computational morphology has the potential to meaningfully contribute to language documentation projects, but only if we center the needs of real documentary linguists through UCD.
\section{On NLP for Language Documentation}
Documentary linguistics aims to create records of human languages through collections of linguistic materials. While not inherently tied to language revitalization, language documentation is often part of broader efforts by marginalized communities to reclaim and strengthen languages impacted by oppression and endangerment. With nearly half of the world’s approximately 7000 languages considered endangered \cite{Bromham_Dinnage_Skirgård_Ritchie_Cardillo_Meakins_Greenhill_Hua_2022}, this work is increasingly urgent.

However, documentation is complex and resource-intensive. It requires linguistic expertise and long-term, collaborative engagement with speakers—especially when aligned with revitalization goals. These efforts must be carefully planned and ethically informed \cite{bird-2024-must, schwartz-2022-primum}. Although there is growing interest in NLP tools to support this endeavor, and many works on computational morphology list supporting language documentation as an explicit aim \citep[\textit{inter alia}]{moeller-hulden-2018-automatic, Moeller_Hulden_2021, liu-etal-2021-morphological,  Moeller_2021, ginn-etal-2023-findings, ginn-etal-2024-glosslm, rice-etal-2024-tams}, widespread adoption remains limited \cite{gessler-2022-closing,gessler-von-der-wense-2024-nlp}. In this section, we examine this tension through the lens of computational morphology as a field positioned to support language documentation.

\subsection{Automated IGT Generation}
Tasks in the area of computational morphology -- such as paradigm completion \cite{kann-schutze-2018-neural}, morphological inflection \cite{cotterell-etal-2016-sigmorphon}, morphological segmentation \cite{kay-1973-morphological,van-den-bosch-daelemans-1999-memory} or morphological tagging \cite{oflazer1994tagging,hajic-hladka-1998-tagging-inflective} -- are frequently motivated by the goal of supporting documentary linguists. Among them,
interlinear glossed text (IGT) generation \cite{ginn-etal-2023-findings} stands out as particularly relevant for language documentation. IGT is a form of morphological annotation that typically adheres to the Leipzig glossing format \cite{Lehmann_1982}, a linguistic representation wherein each line of the target text is broken up into a transcription line, a morphological segmentation line, a gloss line (morphological annotation), and a translation line, though sometimes the transcription is omitted. 
For reference, Example \ref{ex:arapaho-gloss} from \citet{cowell2020} shows an IGT instance in Arapaho, with glosses and translations in English:

\begin{small}
\begin{exe}
  \ex
  \gll nuhu' tih-'eeneti-3i' heneenei3oobei-3i' \\
       this when.PAST-speak-3PL IC.tell.the.truth-3PL \\
  \glt ``When they speak, they tell the truth.''
  \label{ex:arapaho-gloss}
\end{exe}
\end{small}
IGT is a crucial resource for language documentation, but many field recordings fail to progress to IGT because it is expensive and time consuming to create \cite{Seifart_Evans_Hammarström_Levinson_2018}. 

IGT generation is an increasingly popular research area, and promising systems have emerged, ostensibly with the goal of addressing this bottleneck \cite{girrbach-2023-tu-cl, ginn-etal-2024-teach, he-etal-2024-wav2gloss, shandilya-palmer-2025-boosting}. This is thanks in large part to the SIGMORPHON 2023 Shared Task on Interlinear Glossing \cite{ginn-etal-2023-findings}, which provided standard datasets and established an evaluation metric for comparing systems for automated glossing. At the time of writing, the SOTA on five out of seven shared-task languages is held by \textbf{GlossLM} \cite{ginn-etal-2024-glosslm}, a massively multilingual pretrained model for IGT and the subject of our case study in Section \ref{sec:case-study}. We choose to investigate GlossLM not only because of its performance, but also because it is designed to be capable of glossing any language.

\subsection{The NLP gap Revisited}

Despite a growing body of relevant research, and evidence that NLP has the potential to support language documentation \cite{palmer-etal-2009-evaluating, Moeller_Liu_Yang_Kann_Hulden_2020, moeller-hulden-2021-integrating, Chaudhary_2022, ahumada-etal-2022-educational}, NLP systems have not been  widely adopted in documentation workflows \cite{computel-2014-22, flavelle-lachler-2023-strengthening}. \citet{gessler-2022-closing} identifies this disconnect as the "NLP gap," and attribute it to technical limitations, such as poor interoperability between existing NLP tools on the one hand, and the applications used by documentary linguists on the other. Others highlight broader institutional and disciplinary barriers, including conflicting incentives, and limited interdisciplinary training \citep{flavelle-lachler-2023-strengthening}. 

We argue that the NLP gap is compounded by a narrow formulation of research aims, which we see clearly within computational morphology. There have been many shared tasks in areas relevant to the language documentation workflow -- segmentation \cite{batsuren-etal-2022-sigmorphon}, inflection \cite{cotterell-etal-2018-conll, vylomova-etal-2020-sigmorphon, goldman-etal-2023-sigmorphon}, IGT generation \cite{ginn-etal-2023-findings}, and morphosyntactic transformation for the creation of educational materials \cite{chiruzzo-etal-2024-findings, de-gibert-etal-2025-findings}, to name a few. While these tasks have been crucial for driving research and scientific progress, they often rely on simplifying assumptions both in the task formulation and system evaluation. While such simplifications make complex challenges more approachable and help researchers gain traction, they also limit the relevance of research outputs to real-world contexts, resulting in limited adoption by documentary linguists. We argue that it is time to re-assess how we frame tasks, to ensure that our research efforts are strategically directed and that our outputs are as practically useful as intended \cite{kann-etal-2022-major}.
\vspace{-1.75pt}
\section{On Impractical Systems}
\vspace{-1.75pt}
\label{sec:impractical} 
We zoom out for a moment to consider the cultural and epistemic factors that contribute to impractical research outputs from the field of NLP more broadly. Through this lens, we see the NLP gap in language documentation as a product of more systemic challenges. Rethinking the way that we approach usability, particularly in the context of interdisciplinary work, may alleviate some of these long-standing issues.

\subsection{How Our Systems Fail (and How We Fail to Notice)}
Thirteen years ago, \citet{Wagstaff_2012} identified a trend in machine learning research that often paid little heed to real-world impact. She wrote, "This trend has been going on for at least 20 years. Jaime Carbonell, then editor of Machine Learning, wrote in 1992 that 'the standard Irvine data sets are used to determine percent accuracy of concept classification, without regard to performance on a larger external task' \cite{Carbonell_1992}. Can we change that trend for the next 20 years? Do we want to?" \cite{Wagstaff_2012}. Although Wagstaff and Carbonell focused narrowly on classification, similar trends are visible in the field of NLP broadly. 

NLP researchers are still grappling with the shortcomings of our evaluative standard. In a survey of papers published in the \textit{NLP applications} tracks of two major 2020 NLP conferences, \citet{ganesh-etal-2023-mind} found that nearly half lacked evaluations that reflected realistic deployment settings.  If such gaps exist even in the \textit{NLP applications} track—ostensibly focused on systems with practical utility—what does that imply about the field of NLP more broadly? 

This misalignment would be less troubling if our standard metrics were always reliable proxies for downstream utility, but they are not. A growing body of work has shown that intrinsic evaluations often fail to predict real-world effectiveness and/or to align with human preferences \citep[\textit{inter alia}]{ethayarajh-jurafsky-2020-utility, kunz-etal-2022-human, callison-burch-etal-2006-evaluating}. And yet, these metrics continue to dominate how we define and reward success.

\citet{Kogkalidis_Chatzikyriakidis_2024} argue that NLP places disproportionate emphasis on positivist ideals: emphasizing the epistemic value of quantifiable advancements at the expense of social context and theoretical depth. As a result, the field risks becoming increasingly decontextualized from its aims, yielding systems and evaluation practices detached from societal grounding. There are few incentives or standards to ground work in its real-world impact. This is not simply a methodological issue, but a disciplinary one. We are epistemically insular, hesitant to adopt the standards or frameworks of neighboring disciplines, even when tackling problems that clearly demand them \citep{Raji_Scheuerman_Amironesei_2021}. 

These challenges become especially visible in areas like NLP for language documentation, where collaboration with linguists, community stakeholders, and domain experts is essential. Yet, cultural and disciplinary divides have long hindered effective coordination between Indigenous communities, documentary linguists, and NLP researchers \cite{forbes-etal-2022-dim,flavelle-lachler-2023-strengthening, gessler-von-der-wense-2024-nlp}.

If NLP researchers are serious about contributing to language documentation, \textbf{we need evaluation frameworks and design processes that reflect the realities and goals of those we hope to support.} This means actively bridging disciplinary gaps—not just through consultation, but through methods that encourage shared understanding and ongoing dialogue. Without such grounding, our systems risk remaining disconnected from the very communities and contexts they aim to serve.

\subsection{Evaluation as Iteration: User-Centered Design for Improving Research Realism}

Concomitant with an increasing number of NLP researchers acknowledging the limitations of current approaches to usability is a growing movement to restructure research workflows by incorporating methodologies from HCI. In their tutorial on Human-Centered Evaluation of Language Technologies, \citet{Blodgett_Cheung_Liao_Xiao_2024} emphasize that "HCI researchers have developed a 'toolbox of methods' as different 'ways of knowing' \cite{Olson_Kellogg_2014} people’s needs, usage, and interaction outcomes with technologies." Rather than reinventing the wheel, NLP researchers can draw on this body of work to better design systems that are attuned to real-world contexts and user needs.

One such method is User-Centered Design (UCD)—a framework that places user experience at the core of system development through iterative cycles of design, prototyping, and feedback \cite{ergonomics, ucd}. \textbf{UCD encourages researchers to engage with users early and often, integrating their needs, constraints, and environments into every stage of the research and development process.} In doing so, it helps ensure that systems are not only technically effective but also accessible, relevant, and usable in practice.

We are not the first to propose integrating UCD into NLP for linguistic applications \citep[\textit{inter alia}]{adler-etal-2024-user, lyding-schone-2016-design, Ogden_Bernick_1996}, but we argue that it has not been effectively applied to computational morphology, and this blind spot is detrimental to our field's ability to contribute meaningfully to language documentation. UCD is especially necessary in interdisciplinary domains where researchers must navigate the varied perspectives of diverse stakeholders. These contexts are complex by definition, and it is rarely possible to grasp their full nuance without active input from all parties involved. UCD offers a structure for collaboration that supports grounded and reciprocal communication. By foregrounding iterative design and concrete prototypes, UCD helps shift discussions from abstract expectations to tangible possibilities. This framing is especially powerful for communicating across disciplines, where it is challenging to articulate what NLP methods can and cannot do. Presenting early-stage artifacts enables more productive dialogue by anchoring conversations in shared reference points.  

\subsection{More Useful = More Interesting}
It is readily apparent how UCD fits into engineering as a discipline; the goal of engineering is to develop effective systems that support human needs, so centering the user is intuitive. It may be less apparent how user-centered design fits into research, where our goals are more abstract. However, a simple mindset shift illuminates the potential synergy between UCD and research.

Interdisciplinary problems have inherently complex, multi-dimensional solution spaces. When we divorce our research from real-world context, we construct simulacra—crude approximations of reality that lack depth and nuance. In doing so, we risk losing the little details that make problems meaningful— details that could become footholds for future work.  \textbf{In recontextualizing NLP through UCD, we open the door for novel research directions that are not only more useful but more interesting as well.}
\section{User-Centered Design for Automatic Interlinear Glossed Text Generation: A GlossLM Case Study}
\label{sec:case-study}
We describe a case study on the usability of GlossLM -- a multilingual pretrained IGT generation model-- for documentary linguistics. We treat GlossLM as an assistive glossing tool, intended to slot into existing documentation workflows and supplement human annotation efforts. 
Early work in active learning for morpheme glossing \cite{baldridge-palmer-2009-well} shows that the strategy of a documentary linguist correcting machine label suggestions is faster than that same linguist labeling everything manually from scratch.
We recruit 3 expert linguists to complete a small annotation task in their respective languages of expertise -- Teotitlán del Valle Zapotec, Kotiria, and Arapaho -- and interrogate their experience through surveys and interviews.\footnote{This work was approved by our institution's board for responsible research.}

The study was originally conceived as a traditional user study—a post-hoc evaluation rather than part of the development process. However, in interacting with linguists, we encounter several concrete limitations of GlossLM that shift our perspective, provoking critical research questions and revealing promising, research-driven extensions to the system and underscoring the value of user-centered design as an iterative process. Our focused, small-scale interview process yields rich insights, demonstrating that even lightweight, early-stage engagement could meaningfully shape system development.

\subsection{GlossLM Model Details}
GlossLM is a ByT5 \cite{Raffel_Shazeer_Roberts_Lee_Narang_Matena_Zhou_Li_Liu_2020} model continually pretrained on 450k IGT instances spanning 1,800 languages. Leveraging effective crosslingual transfer, GlossLM can accurately generate glosses for a wide range of languages, making it a promising solution for low-resource scenarios where training monolingual models is not feasible. Notably, it achieves state-of-the-art performance on five of seven languages in the SIGMORPHON shared task -- including Arapaho -- highlighting a valuable opportunity to examine which aspects of model performance are not fully captured by standard evaluation metrics.

\subsection{Annotators and Annotation Procedures}
We recruit three linguists with 10+ years of experience glossing in Zapotec, Kotiria, and Arapaho.\footnote{Subjects were recruited via personal communication and participated on a volunteer basis. Consent for data use was obtained via an IRB exempted consent form.} We refer to these participants as Linguists Z, K, and A to maintain anonymity. We ask each linguist to provide a corpus consisting of 25 sentences/lines in their language of study with corresponding English translations. We process this data with GlossLM, passing the target language transcription and English translation as input to the model. We then return the GlossLM outputs to each linguist and request that they manually correct the generated glosses. We do not give strict glossing guidelines, asking instead that they attempt to simulate their preferred glossing conventions. Our aim is to discern whether GlossLM effectively supports a range of glossing habits, as IGT standards vary drastically from person to person \cite{Chelliah_Burke_Heaton_2021}. Following their completion of the annotation task, we ask that each participant respond to a survey and sit for a 30-minute interview.  

\subsection{Survey}
We design our survey to capture initial impressions from our participants immediately after completing the annotation task. We ask 8 questions concerning the ease, accuracy, and efficiency of correcting GlossLM generated glosses. The questions are provided in Appendix \ref{appendix:survey}.

\begin{table*}[ht]
\centering
\begin{tabular}{lccc}
\hline
\textbf{Language} & \textbf{chrF++} & \textbf{\% of Pretraining Corpus} & \textbf{\# of Pretraining Samples}\\
\hline
Teotitlán del Valle Zapotec  &  \texttt{--} & \texttt{0.00826} & \texttt{28} \\
Kotiria &  \texttt{15.04} & \texttt{0.0876} & \texttt{297}\\
Arapaho  &  \texttt{79.45} & \texttt{10.9} & \texttt{36957}\\
\hline
\end{tabular}
\caption{chrF++ \cite{popovic-2015-chrf} scores of GlossLM on task corpus and proportional representation in pretraining data for Zapotec, Kotiria, and Arapaho. \textit{Note: we do not have gold glosses for Zapotec so we do not compute chrF++.}}
\label{tab:lang_stats}
\end{table*}

\subsection{Interview}
In addition to our survey, we conduct, record, and transcribe 30-minute open-ended interviews with each of our participants. The goal of the interviews is to attain more thorough and nuanced perspectives on participants' experience with GlossLM and more general thoughts about the role of NLP in linguistic documentation. While our specific inquiries are context-dependent and vary between interviewees, our guiding questions are as follows: (1) \textit{Describe your usual process for working with your collected data, and especially for glossing.}
(2) \textit{Did you notice any patterns (anything interesting?) in the mistakes that GlossLM made, or in the things that it did well?}
(3) \textit{Is there anything you would change about our strategy for incorporating GlossLM outputs? If yes, how would your suggested configuration better aid your annotation experience?}
(4) \textit{In this study, we have focused on morpheme glossing. Are there other parts of the documentation workflow where you think support from automated tools would be especially helpful?}
(5) \textit{What are your thoughts about artificial intelligence and its role in linguistics?}

\subsection{Results}
To contextualize our findings, Table \ref{tab:lang_stats} presents the chrF++ \cite{popovic-2015-chrf} scores of GlossLM on each task corpus, alongside statistics reflecting each language’s representation in the GlossLM pretraining data. Our three subject languages sit at three different points along the continuum: Zapotec is nearly unrepresented in the pretraining corpus, Kotiria is close to the amount of pretraining we would see if the corpus was equally distributed over all 1800 languages, and Arapaho is disproportionately well-represented.

\subsubsection{Survey}
Survey respondents answer several questions unanimously across the board. When asked if glossing conventions in GlossLM matched what they were expecting, respondents answer "somewhat." Prompted to elaborate, participants identify issues with extraneous labels and inaccurate tags on multimorphemic words. Participants also agree that annotating their texts from scratch would be both easier and faster than correcting the GlossLM outputs. Notably, this includes Linguist A--despite GlossLM’s strong performance on Arapaho--who cites problems with alignment and segmentation. We interrogate these concerns more thoroughly in our follow-up interviews.

\subsubsection{Interview}
Through our interviews, we identify four key weaknesses, raising important conceptual questions that we consider avenues for future work.

\paragraph{Can (and should) we constrain glossing model outputs to pre-defined language specific labels? Or should we instead standardize glossing labels across languages?}

Two participants note that GlossLM tends to generate glosses that are not appropriate in the target language. Linguist Z shares, "[In the GlossLM outputs,] verbs were already indicated for third person in some cases. But [in] Zapotec either you have noun phrase or an enclitic, then it gets the third person. So the third person is not incorporated as part of the verb meaning." Similarly, Linguist K notices that there "seem to be some assumptions that you've got person prefixes which don't exist in Kotiria." 

Given that Zapotec and Kotiria make up a relatively small percentage of the model's pretraining data (see Table \ref{tab:lang_stats}), it is unsurprising that GlossLM would be bad at generalizing about their prefixal morphology, but what is notable here is the pattern of mistakes. GlossLM seems to repeatedly make the same/similar errant assumptions about the morphology of the target languages. 
It is highly probable in these instances that GlossLM is generating glosses that are aligned more closely with some other language in its pretraining data.\footnote{It would be interesting to analyze this phenomenon more concretely-- searching the pretraining data to determine whether the offending labels actually exist and which languages they are associated with. The kind of error annotation we would need for this kind of analysis was not part of the original task posed to the annotators.}  These kinds of errors are an inherent pitfall of multilingual models: the tendency to overgeneralize to high-resource or overrepresented languages \cite{wu-dredze-2020-languages}. This begs the question: should we somehow constrain glossing model outputs to language-specific labels?

\paragraph{Can (and should) we tune glossing model outputs to fit the personal glossing conventions of individual linguists?}
In a related issue, the same two participants state that the glossing conventions reflected in the GlossLM outputs did not always match what they were expecting. "[GlossLM] just invents lots of glosses," said Linguist K,  "I don't know what some of them are supposed to mean, like NARR, I'm not sure what that's supposed to mean." In the same vein, Linguist Z mentions that they do not personally use many of the labels that GlossLM output. It is possible that the offending labels were hallucinated, but--since glossing conventions vary even between linguists studying the same language--it is also possible that they were at least somewhat appropriate.\footnote{None of the glosses labeled as Kotiria contain "NARR" in the pretraining data. However, there are 314 occurrences of "NARR" in pretraining instances labeled "Unknown language", so it possible, though improbable, that there is some instance of Kotiria glossed with "NARR" in the pretraining corpus.} Regardless, this finding raises some broader questions about automatic IGT generation: If the subspace of potential glossing standards is theoretically infinite, how can we generate glosses that align with the expectations of individual linguists? Do we need to? 

\paragraph{Can we do accurate glossing without incorporating declarative language-specific information?} 
Some of our participants note the systematicity of some of the performance issues raised above (e.g., misuse of person prefixes in Kotiria) and suggests that these issues could be mitigated if the system could be given a few language-specific rules to steer its outputs. Linguist K suggests that it might help to manually annotate a set of a dozen of the most common grammatical morphemes and let these influence GlossLM's outputs. All three reference Toolbox,\footnote{https://software.sil.org/toolbox/} a language data management software that (among many other functions) suggests morpheme segmentation and glossing for words based on its existing database for the language. 
This functionality is useful to ease the workload of repetitive glossing, but it relies on simple lookup and lacks capacity to generalize to new inputs.

The errors seen in our small samples for each language already show enough regularity to be partially correctable through the application of declarative knowledge about the language, in the form of general language-specific constraints (e.g., ``Kotiria does not use person prefixes on verbs'') or specific tag-label associations (e.g., ``The morpheme X in Kotiria should be labeled as either PST or COMP'').
This finding suggests the potential value of pursuing two different research directions: use of hybrid systems incorporating linguistic resources into neural glossing architectures \cite[][\textit{inter alia}]{mcmillan-major-2020-automating,zhang-etal-2024-hire,yang-etal-2024-multiple}, perhaps as a second layer over outputs from multilingual pretrained models; and use of human-in-the-loop strategies \cite[][\textit{inter alia}]{muradoglu-hulden-2022-eeny,moeller-arppe-2024-machine}.

\paragraph{Can we extract latent segmentation from glossing models? Do we need to?}
All three of our participants agree on a key weakness that makes GlossLM unsuitable for practical applications: lack of morphemic segmentation. Linguist A specifically points to the lack of segmentation as the primary reason that they would not use GlossLM in spite of the model's ostensibly high performance on Arapaho.

Typically, in language documentation, segmentation is done before or in parallel with glossing because IGT relies on morpheme-by-morpheme correspondence. Linguists often gloss by referring back and forth between the segmentation and gloss lines.
GlossLM, however, generates \textit{only the gloss line} so our participants experience the task as a convoluted workflow which expects them to reverse-engineer the segmentation from the gloss. This process likely results in a higher cognitive load than glossing from scratch.

Thus, the outputs of GlossLM fundamentally do not match the ways that linguists interact with data while glossing. We suspect this mismatch comes from the very sensible engineering decision of aligning GlossLM's outputs with the evaluation format required by the shared task on interlinear glossing (the GlossLM paper evaluates on the test data from the shared task). The shared task evaluation, in turn, offers a more attainable task setting than the full segmentation-plus-glossing process.

GlossLM offers a setting in which the model glosses pre-segmented text, but this does not necessarily map to a real-world scenario, since it would be unusual for a linguist to have an unglossed but gold-segmented corpus. Linguists do not typically segment and then gloss whole texts in sequence but instead segment and gloss in parallel on an sentence-by-sentence basis.

Another option would be to pair GlossLM with a separate segmentation model in a cascaded approach, first segmenting a corpus and then passing the output into GlossLM. This may be viable, but it relies on the availability of an effective segmentation model. Chaining two models may also result in propagation of error and worse glossing outputs overall. For example, \citet{he-etal-2024-wav2gloss} investigate both end-to-end models and cascaded pipelines for language documentation tasks and show that pipeline models perform worse on glossing than both single task and multi-task models.

And after all, why should we need a separate segmentation model? Glossing implicitly relies on segmentation, as the labels must correspond to morphemes. Accessing and exposing the model's internal latent segmentation seems a natural next step for addressing the mismatch between model outputs and user needs.

\subsection{Discussion}
We analyze the results of our case-study with reference to several key points from §\ref{sec:impractical}.
\paragraph{We need evaluation frameworks and design processes that reflect the realities and goals of those we hope to support.}
Prior to this study, GlossLM had only been evaluated according to standard metrics specified by the SIGMORPHON 2023 Shared Task on Interlinear Glossing. Its efficacy was reported in abstract with respect to a wide range of languages. This is not necessarily negative -- standardization and abstraction enable straightforward evaluation and easier model comparison. However, the results of our case-study reveal that, despite achieving SOTA on shared-task metrics, GlossLM is not useful to its intended end-users in its current state. This finding supports the notion that metrics are not always a good proxy for downstream utility, and that they should be viewed as part of a larger picture. User studies enable us to put metrics in context and evaluate our systems holistically with respect to downstream realities.

\paragraph{UCD encourages researchers to engage with users early and often, integrating their needs, constraints, and environments into every stage of the research and development process.}
A direct extension from the previous point is that UCD enables researchers and developers to discover and meaningfully address real-world system weaknesses. Our case-study reveals several shortcomings of GlossLM which could have been identified earlier if UCD had been integrated into the initial development process. Our findings underscore the point that system design ought to be iterative, and researchers can and should engage with users to identify and respond to real-world needs.

An important aspect of what we learn from this case study is that invaluable insights can come from working with even a single user, if the user is able to interact with system outputs and share their insights early in the research and development process.

\paragraph{In recontextualizing NLP through user-centered design, we open the door for novel research directions.}
Through this case study, we identify  several  interesting research directions that could yield viable extensions to GlossLM.
\begin{itemize}
    \item[\textbf{Q1.}] Can (and should) we constrain glossing model outputs to pre-defined language specific labels? Or should we instead standardize glossing labels across languages?
    \item[\textbf{Q2.}] Can (and should) we tune glossing model outputs to fit the personal glossing conventions of individual linguists?
    \item[\textbf{Q3.}] Can we do accurate glossing without incorporating declarative language-specific
    information?
    \item[\textbf{Q4.}] Can we extract latent segmentation from glossing models? Do we need to?
\end{itemize}
While a domain expert could certainly come up with these questions independently, grounding them in user studies verifies that they represent research directions that support meaningful contributions to real-world applications.

Our case study illustrates the potential for UCD to be mutually beneficial: addressing the real-world needs of documentary linguists while simultaneously driving novel research contributions in NLP. After sharing these insights with the researchers behind GlossLM, they have embarked on a next iteration: incorporating segmentation into the system outputs.

\section{Conclusion}
The disconnect between NLP research and the realities of language documentation has been repeatedly diagnosed but insufficiently addressed. Within NLP, tasks in computational morphology are especially relevant for the workflow of documentary linguists. We argue that the "NLP gap" in language documentation is a symptom of a broader misalignment between research and practice in NLP--one that we must address, especially because we work with and impact vulnerable communities.

Our case study on GlossLM offers an example of how principles from User-Centered Design (UCD) can meaningfully reshape computational morphology research. We interview three linguists about their experiences with GlossLM, a state-of-the-art model for interlinear glossed text generation, and find that despite impressive performance by standard metrics, the model is unusable in practice. Lack of segmentation, mismatched glossing conventions, and poorly suited label inventories make it difficult to integrate into real documentation workflows. Crucially, these conversations surface more than just critique--they clarify user requirements, offer insight into domain-specific needs, and open new directions for future research.

Closing the NLP gap in language documentation will require more than state-of-the-art models. We will need usable software, sustained collaborations, and careful attention to context and usability. We hope this work serves as both a call to action and a proof of concept—demonstrating that even small focused efforts toward user-centered NLP can generate meaningful findings.

\label{sec:bibtex}

\section*{Limitations}
Our case-study has limited generalizabilty because it consists of only three languages/participants reporting feedback on a single tool. We also acknowledge that qualitative evaluation is inherently subjective and only tells part of the story. A formal user study with quantitative measures of efficiency would be beneficial and complementary.

The study should not be taken as a comprehensive review--it is instead intended to inspire future work on UCD for computational morphology. There are far more insights to be gleaned from interacting with more linguists and experimenting with novel tools on a variety of languages.

\section*{Acknowledgments}
Parts of this work were supported by the National Science Foundation under Grant No. 2149404, “CAREER: From One Language to Another.” Any opinions, findings, and conclusions or recommendations expressed in this material are those of the authors and do not necessarily reflect the views of the National Science Foundation. This work utilized the Blanca condo computing resource at the University of Colorado Boulder. Blanca is jointly funded by computing users and the University of Colorado Boulder.

\bibliography{anthology,custom}
\bibliographystyle{acl_natbib}

\appendix

\section{Survey Questions}
\label{appendix:survey}
\begin{itemize}
\item Did the glossing conventions in GlossLM match what you were expecting?
    \begin{itemize}
         \item Yes/Somewhat/No
         \item Optional: Free Response
    \end{itemize}
\item Did you find the GlossLM generated IGT to be accurate?
    \begin{itemize}
         \item Mostly inaccurate/Somewhat Inaccurate/Somewhat Accurate/Mostly Accurate
    \end{itemize}
\item How easy/difficult did you find it to correct errors in the GlossLM generations?
    \begin{itemize}
         \item Easy/Somewhat Easy/Neutral/Somewhat Difficult/Difficult
    \end{itemize}
\item Given the options of annotating this text from scratch or using GlossLM, which do you think would be faster?
    \begin{itemize}
         \item From Scratch/With GlossLM
    \end{itemize}
\item Given the options of annotating this text from scratch or using GlossLM, which do you think would be easier
    \begin{itemize}
         \item From Scratch/GlossLM
    \end{itemize}
\item Would you incorporate GlossLM into your workflow going forward?
    \begin{itemize}
         \item Yes/Maybe/No
         \item Optional: Free Response
    \end{itemize}
\item Is there anything that would have made the experience more seamless?
    \begin{itemize}
         \item Free Response
    \end{itemize}
\item Is there anything else you would like to say about the experience?
    \begin{itemize}
         \item Free Response
    \end{itemize}
\end{itemize}

\end{document}